\documentclass{article}

\usepackage[preprint]{neurips_2025} 

\usepackage{amsmath, amssymb}
\usepackage{graphicx}
\usepackage{geometry}
\usepackage{longtable}
\usepackage{array}

\usepackage{booktabs}
\usepackage{hyperref}
\usepackage{caption}
\usepackage{subcaption}
\usepackage{natbib}
\usepackage{xcolor}
\usepackage{bm}
\usepackage{enumitem}

\title{Enhancing Answer Reliability Through Inter-Model Consensus of Large Language Models}

\author{
  Alireza Amiri-Margavi \\
  \texttt{ala170@pitt.edu} \\
  \And
  Iman Jebellat \\
  \texttt{iman.jebellat@mail.mcgill.ca} \\
  \AND
  Ehsan Jebellat \\
  \texttt{jebellat@mines.edu} \\
  \And
  Seyed Pouyan Mousavi Davoudi \\
  \texttt{spouyan.mousavi@gmail.com}
}

\begin{document}

\maketitle

\begin{abstract}
We propose a collaborative framework in which multiple large language models---including GPT-4-0125-preview, Meta-LLaMA-3-70B-Instruct, Claude-3-Opus, and Gemini-1.5-Flash---generate and answer complex, PhD-level statistical questions when definitive ground truth is unavailable. Our study examines how inter-model consensus improves both response reliability and identifies the quality of the generated questions. Employing chi-square tests, Fleiss' Kappa, and confidence interval analysis, we quantify consensus rates and inter-rater agreement to assess both response precision and question quality. Key results indicate that Claude and GPT-4 produce well-structured, less ambiguous questions with a higher inter-rater agreement, as shown by narrower confidence intervals and greater alignment with question-generating models. In contrast, Gemini and LLaMA exhibit greater variability and lower reliability in question formulation. These findings demonstrate that collaborative interactions among large language models enhance response reliability and provide valuable insights for optimizing AI-driven collaborative reasoning systems.

\end{abstract}


\section{Introduction}

The emergence of large language models (LLMs) has transformed natural language processing and artificial intelligence (AI), enabling machines to perform complex language tasks with unprecedented proficiency. Models such as OpenAI's GPT-4, Meta's LLaMA series, Anthropic's Claude,  Google's Gemini, and DeepSeek have demonstrated remarkable capabilities in generating human-like text, understanding context, and even exhibiting reasoning abilities \cite{mann2020language, touvron2023LLaMA, guo2025deepseek}. These advancements have opened new avenues for automated knowledge generation and validation in specialized domains \cite{kojima2022large}.

However, two significant challenges remain: validating model outputs when predefined correct answers or ground truths are unavailable, and assessing the quality of the questions these models generate. These issues are especially critical in specialized fields like advanced statistics, where the complexity of questions makes manual verification resource-intensive and impractical \cite{jiao2023automatic}. In such contexts, traditional validation methods that compare outputs against predetermined answers are simply insufficient \cite{hendrycks2020measuring}.

Leveraging multiple LLMs to enhance the reliability of the answer and assess the quality of the questions through collaboration and consensus presents a promising approach. By utilizing the collective intelligence of several models, it becomes possible to approximate correctness and identify consensus even in the absence of ground truth \cite{he2022wisdom}. Additionally, this strategy provides insights into question quality by evaluating how consistently other models respond. This methodology draws inspiration from ensemble learning techniques in machine learning \cite{dietterich2000ensemble} and the wisdom of crowds phenomenon in human collective intelligence \cite{mennis2006wisdom}.

Recent studies indicate that different LLMs may exhibit complementary strengths and weaknesses in their reasoning capabilities \cite{taylor2022galactica}. This observation suggests that a collaborative framework leveraging multiple models could overcome individual limitations and yield more reliable answers. However, the dynamics of such inter-model collaboration, especially in specialized domains like statistics, remain largely unexplored.

This research synthesizes three fundamental theoretical frameworks: collective intelligence theory, distributed cognition framework, and consensus formation models to understand and analyze the collaborative dynamics among LLMs in statistical reasoning tasks.

Collective Intelligence Theory \cite{levy1997collective} provides a foundational basis for understanding how multiple agents can collaborate to achieve superior outcomes compared to individual performance. In the context of LLMs, this theory suggests that different LLM architectures (e.g., GPT-4, Claude-3, LLaMA, Gemini) offer varied approaches to problem-solving, potentially leading to more robust solutions \cite{hong2004groups}. It informs our approach to combining individual model outputs through structured consensus formation \cite{woolley2010evidence}. Additionally, each model's independent processing of questions helps maintain solution diversity and reduces cascading errors \cite{vercammen2019collective}.

The Distributed Cognition Framework \cite{hutchins1995cognition} offers theoretical grounding for understanding how cognitive processes can be distributed across multiple artificial agents. This framework is particularly relevant because different models may encode complementary aspects of statistical knowledge, leading to more comprehensive problem-solving capabilities \cite{du2024large}. Complex statistical reasoning tasks can be decomposed and processed in multiple models, potentially improving the overall quality of the solution \cite{liu2024large,naik2024probabilistic}. 

Mathematical models of consensus formation provide the theoretical basis for understanding how agreement emerges among multiple decision-making agents \cite{friedkin1990social}. These models inform the fundamental dynamics of collaborative AI systems through three key aspects: the dynamics of agreement, which describe how consensus evolves; weighted influence mechanisms, which define the relative impact of each agent in the consensus process \cite{olfati2007consensus}; and the conditions for convergence that ensure the stability of collaborative solutions \cite{baronchelli2018emergence}. Work by \cite{bahrami2010optimally} has demonstrated how these theoretical frameworks can be effectively applied to multi-agent decision-making systems, offering a robust foundation for understanding collaborative behavior in AI.

Integrating collective intelligence, distributed cognition, and consensus formation frameworks provides a comprehensive foundation for analyzing collaborative AI systems \cite{hutchins1995cognition, woolley2010evidence}. This integration can be formally expressed as:
\begin{equation}
R = f(\text{CI}, \text{DC}, \text{CF})
\end{equation}
where \( R \) represents the reliability of collaborative outcomes, integrated across Collective Intelligence (CI), Distributed Cognition (DC), and Consensus Formation (CF) principles \cite{patrikalakis1999distributed}. Through this unified approach, we can better understand how multiple AI models collaborate and generate reliable solutions in complex reasoning tasks.

Our analytical framework synthesizes quantitative and qualitative methods derived from these theoretical perspectives \cite{klein2006making}. The quantitative aspects draw from consensus formation models, providing metrics for measuring agreement and convergence in multi-agent systems \cite{mennis2006wisdom}. 

The validation methodology emerging from this integrated framework emphasizes the importance of multiple perspectives in evaluating model outputs \cite{kittur2008harnessing}. By considering various theoretical viewpoints, we can better assess the reliability and robustness of collaborative AI solutions \cite{malone2022handbook}. This comprehensive theoretical foundation supports our investigation of established principles and provides new insights into the unique challenges posed by AI collaboration in statistical reasoning tasks \cite{page2008difference}.

Although existing research has extensively examined individual LLM performance \cite{ahn2024large} and basic ensemble methods \cite{huang2024enabling}, there is a notable gap in understanding how multiple state-of-the-art LLMs can collaboratively validate complex knowledge in the absence of ground truth. This gap is particularly significant in specialized academic domains where the complexity of questions demands sophisticated reasoning capabilities, manual validation of both question quality and responses by human experts is time-consuming and costly, and traditional automated validation methods are insufficient. Furthermore, the dynamic nature of knowledge makes maintaining up-to-date ground truths a formidable challenge. Addressing these issues requires leveraging collaborative validation among LLMs, which provides a pathway toward reliable, scalable, and efficient knowledge validation.

This study aims to explore and understand the collaborative dynamics among different LLMs in statistical reasoning tasks.  Our primary objective is to examine how answers from various LLMs align or differ when responding to complex statistical questions generated by one of the models, thereby revealing how these models complement each other and offering insights into the quality of their question-generation capabilities. We further develop a framework for evaluating answer reliability based on model consensus in the absence of a predefined ground truth. To achieve this, we quantify the reliability of collaborative outcomes and identify metrics that best capture inter-model agreement, employing statistical techniques such as chi-square tests and Fleiss' Kappa to assess the significance of consensus rates. These efforts support validating collaborative mechanisms and the effectiveness of consensus formation among multiple LLMs. 

The findings of this study have the potential to influence educational technology, automated assessment, research validation in specialized fields, and the development of more robust AI-powered knowledge systems while enhancing our understanding of collaborative AI.

\section{Literature Review}

AI trends have moved from advances in image processing and computer vision \cite{voulodimos2018deep, azad2024advances} to progress in reinforcement learning \cite{wiering2012reinforcement, jebellat2024reinforcement} and natural language processing \cite{chowdhary2020natural, pillai2023advancements}, leading to today's focus on LLMs and generative AI.

LLMs have achieved significant milestones in natural language understanding and generation \cite{radford2019language}. The GPT series, notably, has demonstrated remarkable capabilities in few-shot and zero-shot learning scenarios \cite{achiam2023gpt}. Recent advancements in model architectures and training approaches have led to increasingly sophisticated systems, such as Claude-3 \cite{anthropic2024claude3}, Gemini \cite{team2023gemini} and DeepSeek \cite{guo2025deepseek}, which exhibit enhanced reasoning abilities in specialized domains.

The concept of collaborative intelligence among AI agents has emerged as a promising approach for enhancing problem-solving capabilities \cite{dafoe2020open}. Recent studies have shown that model collaboration can significantly improve reasoning capabilities through cross-model validation \cite{yin2023exchange}. Additionally, it enhances robustness, particularly in tackling complex problem-solving tasks \cite{sun2023corex, davoodi2024llms}, and leads to more reliable outputs by leveraging consensus mechanisms.

Integrating collaborative strategies in AI systems has consistently demonstrated performance enhancement across various domains. For instance, \cite{lu2024merge} conducted a comprehensive survey highlighting that merging, ensembling, and cooperative approaches among LLMs lead to superior outcomes in natural language processing tasks. In the medical field, \cite{rezk2024metaheuristic} reviewed metaheuristic-based ensemble learning methods, emphasizing their effectiveness in improving diagnostic accuracy and treatment planning. Similarly, in mathematical problem-solving, researchers at MIT developed a multi-AI collaboration framework that enhances reasoning and factual accuracy in LLMs, resulting in more reliable solutions to complex mathematical queries \cite{mit2023multi}. In software engineering, ensemble methods have been shown to improve code defect detection and software quality assurance processes \cite{gupta2022review}. These findings collectively highlight the significant advantages of collaborative AI approaches, particularly as LLMs continue to evolve and scale.

The collaboration of AI models raises critical ethical questions regarding transparency, accountability, and bias propagation \cite{bender2021dangers}. Ensuring that collaborative AI systems operate ethically is crucial, especially in the absence of ground-truth verification \cite{mittelstadt2019principles}. 

A primary concern is the potential for bias amplification in collaborative systems. \cite{raghavan2020mitigating} demonstrated that model ensembles can compound existing biases, with their study showing significant increases in gender and demographic biases when multiple models interact. Building on this work, \cite{hashimoto2018fairness} established frameworks for measuring and mitigating such cumulative biases in machine learning systems.

Transparency presents another significant challenge in collaborative AI systems. \cite{doshi2017accountability} identified critical areas, including decision attribution complexity, interpretability of consensus mechanisms, and accountability frameworks. These challenges become particularly essential in high-stakes applications, where understanding model decisions is crucial \cite{rudin2019stop}.

Recent case studies have highlighted these ethical considerations across various domains. In healthcare applications, \cite{larrazabal2020gender} examined ethical implications in medical imaging diagnosis, finding that collaborative systems required additional safeguards for patient privacy and decision transparency. Similarly, in educational contexts, \cite{holstein2019improving} investigated fairness in automated assessment systems, revealing the need to carefully calibrate collaborative systems to ensure equitable evaluation across diverse student demographics. \cite{gebru2021datasheets} further emphasized the importance of comprehensive documentation and transparency in AI systems, mainly when multiple models work in conjunction.

\section{Methodology}

This study employs a mixed-methods approach to investigate the collaborative dynamics among LLMs in statistical reasoning tasks. The experimental design incorporates a quantitative analysis of model consensus patterns. The study framework can be formalized as follows:

\begin{equation}
S = \{ M, Q, A, V \},
\end{equation}

where \( M \) represents the set of models, \( Q \) the question generation process, \( A \) the answering process, and \( V \) the validation mechanisms.

We examine how multiple LLMs collaborate to generate and validate complex, PhD-level statistical multiple-choice questions (MCQs) without ground-truth answers, while also assessing each LLM's ability to produce meaningful questions. A total of \( N = 100 \) MCQs were generated and answered. An LLM acted as the question generator for each question, while the remaining three independently provided answers and justifications. Question generator and answerer roles were rotated among the models to ensure interchangeability and mitigate model-specific biases.

We utilized four state-of-the-art LLMs, each with distinct architectural characteristics.

\begin{table}[!ht]
\centering
\caption{Descriptions of the Four Language Models Used}
\label{tab:llm_descriptions}
{\small
\begin{tabular}{@{}p{3.5cm}p{10cm}@{}}
\toprule
\textbf{Model}            & \textbf{Description}                                                                                 \\ 
\midrule

GPT-4-0125-preview        & An advanced model renowned for its reasoning capabilities and sophisticated natural language understanding \cite{openai2023gpt4}. \\ 
Meta-LLaMA-3-70B-Instruct & A 70-billion-parameter model optimized for instruction following, featuring enhanced few-shot learning capabilities \cite{touvron2023LLaMA}. \\ 
Claude-3-Opus             & Designed to provide safe and helpful responses focusing on complex tasks \cite{anthropic2024claude3}. \\ 
Gemini-1.5-Flash          & A multimodal model integrating language and vision \cite{team2023gemini}.                                  \\ \bottomrule
\end{tabular}
}
\end{table}

Each model was accessed via its respective API, with standardized key parameters to ensure comparable responses. All models operated under identical conditions to maintain consistency.

To ensure the creation of challenging and diverse multiple choice questions (MCQs) suitable for PhD-level statistics, we designed an integrated framework that combined question generation and independent answering LLMs. This framework also incorporated strategies to mitigate potential biases in the question generation and answering processes.

The question generation phase leveraged a comprehensive concept map of advanced statistical topics, as illustrated in table \ref{fig.Questions_tree}. This table covers a diverse set of topics \( \mathcal{T} \) and subtopics  \( \mathcal{S} \). For each question, a topic \( t_i \in \mathcal{T} \) and a subtopic \( s_i \in \mathcal{S} \) were randomly selected to ensure uniform coverage and diversity. The LLM which generated questions received a carefully designed prompt \( P_q \):

\begin{quote}
\textit{"Generate a challenging multi-choice question at the PhD level in the field of [\textbf{Topic}], focusing on [\textbf{Specific Concept}]. The question should have four answer options labeled A, B, C, and D, with only one correct answer. Ensure that the question tests deep understanding and critical thinking skills."}
\end{quote}

Here, the placeholders [\textbf{Topic}] and [\textbf{Specific Concept}] were populated based on the selected topic \( t_i \) and subtopic \( s_i \) from the table \ref{fig.Questions_tree}. The generated output included the question \( Q_i \), the set of answers \( \{ A_i, B_i, C_i, D_i \} \), the correct answer \( A_i^c \), and an explanation \( E_i \). In particular, \( A_i^c \) and \( E_i \) were retained for analysis but withheld from the answer models to replicate real-world conditions where ground truth answers are not available.

To mitigate potential biases, we implemented several strategies. First, topic diversity was ensured by including various topics and subtopics, preventing over-representing specific areas. Second, the crafted prompts were neutral and avoided introducing leading language or biases, ensuring no influence on the models' responses in a particular direction. Finally, all generated content was manually reviewed to identify and exclude inappropriate or biased material, thus preserving the integrity and fairness of the dataset.

\begin{table}[h!]
\centering
\caption{Comprehensive overview of statistical topics and their key subtopics. This served as the foundation for question topic selection.}
\label{tab:stat_topics}
{\small
\begin{tabular}{@{}p{4cm}p{10cm}@{}}
\toprule
\textbf{Topic}            & \textbf{Subtopics}                                                                                 \\ 
\midrule

Bayesian inference        & Prior distributions, Posterior updating, Bayes factors, Conjugate priors \\ 

Markov Chain Monte Carlo methods & Metropolis-Hastings algorithm, Gibbs sampling, Convergence diagnostics, Sampling efficiency \\ 

Time series analysis      & ARIMA models, Stationary, Seasonality, Spectral analysis \\ 

Multivariate statistics   & Principal Component Analysis, Factor analysis, Canonical correlations, Multivariate normal distribution \\ 

Hypothesis testing        & Type I and Type II errors, Power analysis, Non-parametric tests, Multiple comparisons correction \\ 

Non-parametric methods    & Kernel density estimation, Bootstrap methods, Spline regression \\ 

Survival analysis         & Cox proportional hazards model, Kaplan-Meier estimator, Censoring, Hazard functions \\ 

Experimental design       & Randomization techniques, Blocking and confounding, Factorial designs, Response surface methodology \\ 

Regression analysis       & Generalized linear models, Heteroscedasticity, Collinearity, Model selection criteria \\ 

Statistical learning theory & Overfitting and underfitting, Regularization techniques, Bias-variance tradeoff, Cross-validation \\ 

\bottomrule
\end{tabular}
}
\label{fig.Questions_tree}
\end{table}

In the answer phase, three distinct LLMs independently attempted the generated questions. Each answering model received the question \( Q_i \) and answer set \( \{ A_i, B_i, C_i, D_i \} \) alongside the following prompt \( P_a \):

\begin{quote}
\textit{"Please read the following Ph.D.-level statistics question and select the most appropriate answer (A, B, C, or D). Provide a detailed justification for your selection, explaining your reasoning and any relevant statistical principles."}
\end{quote}

Each model independently selected an answer \( a_{ij} \) for the question \( Q_i \), where \( j \in \{1,2,3\} \) indexes the response models and provided a justification \( J_{ij} \). The models were isolated from each other, ensuring that no information was exchanged about answers or justifications to prevent bias or collusion. This isolation, combined with the neutral prompts and diverse question topics, further minimized bias in the answering process.


  We performed an inter-model consistency analysis to assess the agreement among the answering LLMs and evaluate the reliability of their consensus. For each question \( Q_i \), we collected and organized data into a comprehensive dataset \( D_i \), defined as:

\begin{equation}
D_i = \{ Q_i, \{ A_i, B_i, C_i, D_i \}, \{ a_{i1}, a_{i2}, a_{i3} \}, \{ J_{i1}, J_{i2}, J_{i3} \} \}
\end{equation}

This data set encompassed the generated questions \( Q_i \), their corresponding multiple choice options \( \{ A_i, B_i, C_i, D_i \} \), the responses of each participating LLM - including selected answers \( a_{ij} \) and justifications \( J_{ij} \) - and metadata such as model identifiers, timestamps, and relevant model-specific parameters. This structured approach facilitated the quantitative and qualitative analysis of reasoning processes and agreement patterns among models \cite{bommasani2021opportunities}.

We analyze the alignment and divergence in the selected answers \( a_{ij} \) to quantify the consistency between models. The degree of consensus was categorized into three levels: full agreement, where all three models selected the same answer; partial agreement, where two models selected the same answer while one differed; and no agreement, where all models selected different answers. This classification provided a robust framework for evaluating agreement levels and understanding the variability in model decision-making, providing insights into their collective and individual reasoning patterns.

Given the absence of ground-truth answers, we employed several mechanisms to validate the models' answers based on inter-model consensus.
\paragraph{Majority Vote, Reliability, and Confidence Interval}
To determine the consensus answer \( A_i^{\text{cons}} \) for a given question \( Q_i \), we used a majority voting mechanism over the responses of multiple LLMs. Formally, the consensus answer is defined as:
\begin{equation}
A_i^{\text{cons}} = \arg\max_{k \in \{A, B, C, D\}} \sum_{j=1}^{3} \delta(a_{ij}, k),
\end{equation}
where \( a_{ij} \) denotes the response provided by the \( j \)-th LLM for question \( Q_i \), and \( \delta(a_{ij}, k) \) is the Kronecker delta function:
\[
\delta(a_{ij}, k) =
\begin{cases}
1, & \text{if } a_{ij} = k, \\
0, & \text{otherwise.}
\end{cases}
\]
This formulation assigns a frequency-based score to each answer option \( k \), with the majority-selected response designated as the consensus answer. Both partial and complete agreements are considered valid and form the basis for subsequent analyses.

In the absence of ground truth labels, we introduced a reliability metric to assess the trustworthiness of the consensus answers. Reliability measures the alignment between the consensus response and the answer provided by the LLM that generated the question. Let \( A_i^{\text{LLM-q}} \) denote the querying LLM's response to question \( Q_i \). The reliability score \( R_i \) is computed as:
\begin{equation}
R_i = 
\begin{cases}
1, & \text{if } A_i^{\text{cons}} = A_i^{\text{LLM-q}}, \\
0, & \text{otherwise.}
\end{cases}
\end{equation}
A higher reliability score indicates a more substantial alignment between the consensus answer and the querying LLM's response, increasing confidence in the consensus's validity. This measure is particularly valuable in settings where ground truth data is unavailable, and, when combined with consensus analysis, it provides qualitative insights into the performance of models in question generation tasks.

To evaluate the robustness of the consensus rates in LLMs, we calculated confidence intervals (CI) using a bootstrap resampling approach. Confidence intervals provide a probabilistic range for the actual consensus rate, capturing variability in the observed data. Narrow CIs indicate high reliability, while wider CIs reflect greater uncertainty.

Bootstrap resampling enables CI estimation without assumptions about the underlying data distribution. Specifically, \( B=1000 \) bootstrap samples are generated by sampling the original data set with replacement. The mean agreement rate is computed for each sample, constructing a bootstrap distribution of the consensus rate. The \( \alpha/2 \)-th and \( 1-\alpha/2 \)-th percentiles of this distribution define the lower and upper CI bounds, respectively. For a 95\% CI, these correspond to the 2.5 and 97.5 percentiles.

By incorporating confidence intervals, we mitigate the uncertainty in agreement rates and establish a statistical basis for comparing LLMs. Overlapping CIs suggest that there are no significant differences between models, whereas nonoverlapping intervals indicate statistically significant differences. This methodology ensures a robust and interpretable evaluation of model reliability, mainly when ground truth labels are unavailable.

To further validate reliability and assess statistical significance, we employed the following additional statistical measures.

\paragraph{Chi-Square Test of Independence}
A chi-square test was used to evaluate whether the distribution of selected responses deviated significantly from random chance. The test statistic was calculated as:

\begin{equation}
\chi^2 = \sum_{k=1}^{K} \frac{(O_k - E_k)^2}{E_k},
\label{chi_square_test}
\end{equation}

Where:

\begin{itemize}
    \item \( O_k \) is the observed frequency of the answer choice \( k \) in all models and questions.
    \item \( E_k \) is the expected frequency of the answer choice \( k \), assuming uniform random selection \( E_k = \frac{N \times n_j}{K} \).
    \item \( K = 4 \) is the number of choice of responses.
    \item \( N \) is the total number of questions.
    \item \( n_j = 3 \) is the number of response models.
\end{itemize}

To determine statistical significance, the computed \( \chi^2 \) value was compared against the chi-square distribution with \( K - 1 \) degrees of freedom.

\paragraph{Fleiss' Kappa Coefficient}

Fleiss' kappa \( \kappa \) was employed to quantify the agreement among the models beyond chance. It is defined as:

\begin{equation}
\kappa = \frac{\overline{P} - \overline{P_e}}{1 - \overline{P_e}},
\end{equation}

Where:

\begin{itemize}
    \item \( \overline{P} \) is the mean observed agreement among the models.
    \item \( \overline{P_e} \) is the mean agreement expected by chance.
\end{itemize}

The value of \( \kappa \) ranges from \( -1 \) (perfect disagreement) to \( 1 \) (perfect agreement), with \( 0 \) indicating that there is no agreement beyond chance.

These mechanisms collectively ensured a robust framework for validating model responses and assessing the reliability of the consensus generated.

\section{Results}

We evaluated the level of agreement among the models—GPT-4, LLaMA, Gemini, and Claude—for each generated question. In each experiment, one model generated $100$ questions, and the remaining three independently provided answers without collaboration. The data are available here \cite{amiri2024intermodel}. 

\subsection*{Inter-Model Consistency Analysis}
\begin{figure}[htbp]
    \centering
    \includegraphics[width=0.8\linewidth]{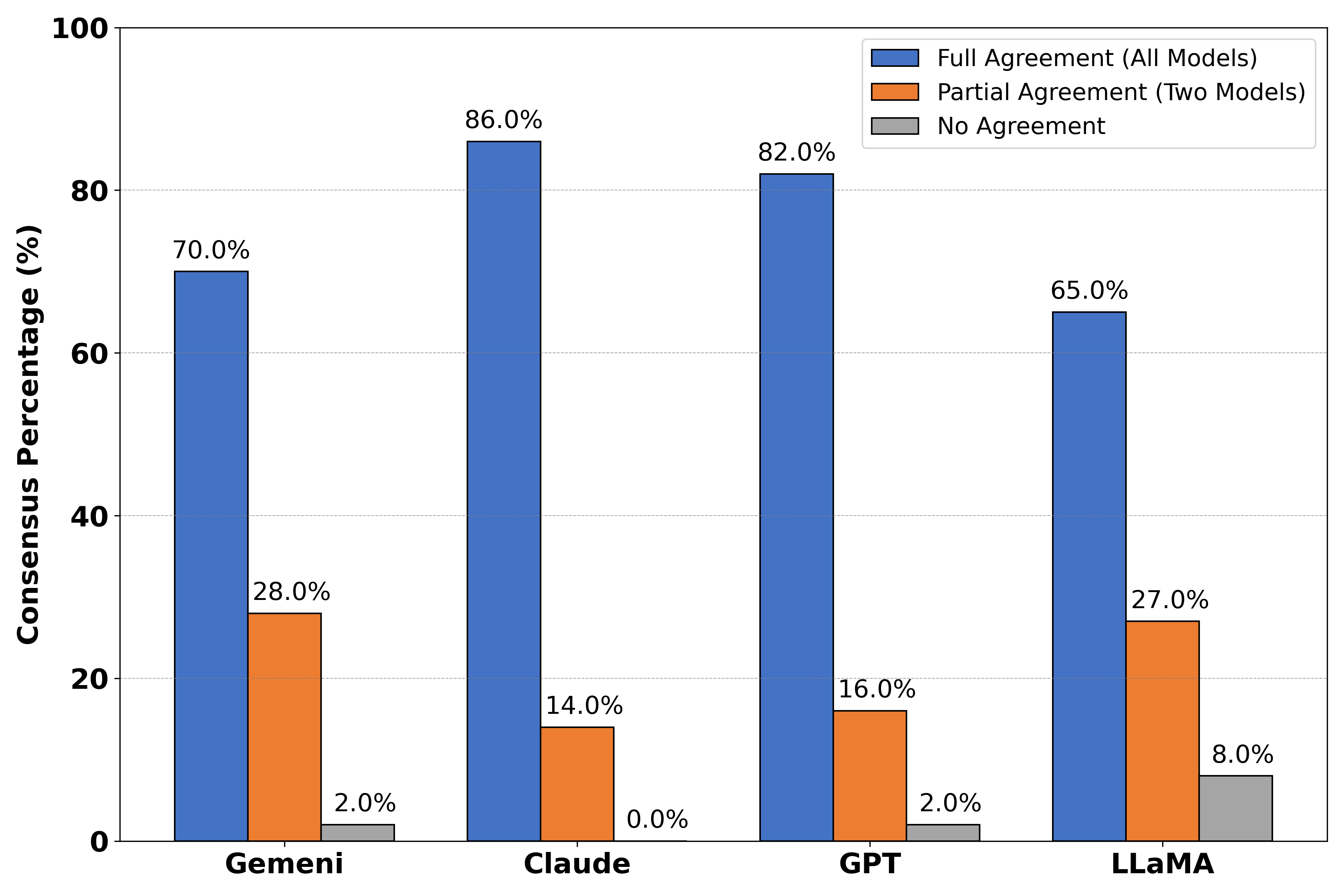}
    \caption{Consensus rate overview showing the levels of agreement among the models when different LLMs generate the questions. The x-axis represents the question-generating LLM model, and the y-axis indicates the percentage of responses in each agreement category.}
    \label{fig:consensus_rate_overview}
\end{figure}

Figure~\ref{fig:consensus_rate_overview} illustrates the levels of agreement between the models based on their responses to the questions generated by different LLMs. The results indicate that Claude and GPT-4 achieved the highest full agreement rates, with $86\%$ and $82\%$ of the responses aligning when they served as the question generator. In contrast, Gemini and LLaMA showed lower consensus, with full agreement rates of $70\%$ and $65\%$, respectively. This suggests that the questions generated by Claude are more comprehensible to other LLMs, as evidenced by the consistent selection of the same answer among multiple choice options \(\{ A, B, C, D \}\). In contrast, when LLaMA generated the questions, only $65\%$ of the responses reached full agreement, indicating greater variability in interpretation.

Further analysis supports these observations. LLaMA-generated questions resulted in a partial agreement rate of $27\%$ and the highest non-agreement rate of $8\%$, suggesting that their questions may be more ambiguous or challenging to interpret. In contrast, Claude's questions had a partial agreement rate of only $14\%$ and no cases of complete disagreement, supporting the hypothesis that their ability to generate well-structured and comprehensible questions was compared to the other LLMs, particularly LLaMA.

 To evaluate the reliability and precision of the answers generated by LLM in scenarios lacking ground-truth data, we used two complementary mechanisms: majority voting and confidence interval analysis. These approaches quantify inter-model agreement and assess alignment with a reference answer provided by the question-generating LLM.

\begin{figure}[htbp]
    \centering
    \includegraphics[width=0.8\linewidth]{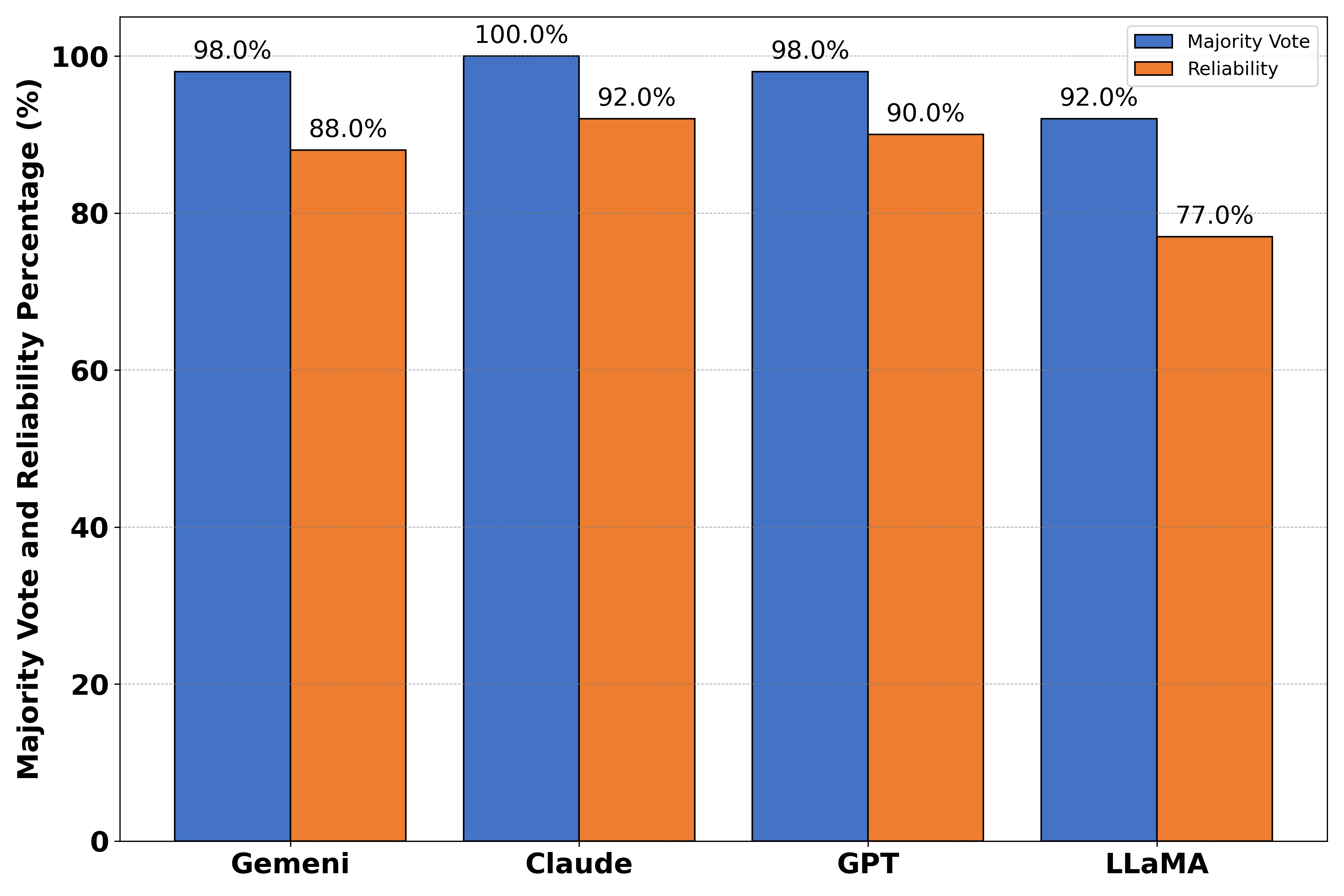}
    \caption{Majority vote and reliability percentages across different models. The majority vote percentage indicates instances where at least two models agree, while the reliability percentage reflects instances where two or more models agree with the question-generating LLM's answer.}
    \label{fig:majority_agreement_reliability}
\end{figure}

Using the majority voting approach, an answer is considered reliable if at least two other LLMs agree with the question-generating LLM's response.  Figure \ref{fig:majority_agreement_reliability} presents the percentages of majority vote and reliability of various LLMs.  The results indicate that the percentages of majority votes remain high in all models, with Claude achieving the highest at $100\%$, followed by Gemini and GPT-4 at $98\%$, and LLaMA at $92\%$. This suggests that in most cases, two or three LLMs consistently selected the same answer among multiple choice options \(\{ A, B, C, D \}\).

Reliability analysis reveals notable differences in the frequency with which models align with the reference answer. Claude demonstrates the highest reliability percentage ($92\%$), followed closely by GPT-4 ($90\%$) and Gemini ($88\%$), suggesting strong consistency with the LLM response generating the questions. In contrast, LLaMA exhibits the lowest reliability percentage ($77\%$), indicating a reduced alignment with reference answers.

To further evaluate the precision and reliability of inter-model agreement, we calculated confidence intervals (CIs) for the consensus rates of each model. Confidence intervals provide a statistical range within which the consensus rate is likely to fall, allowing an assessment of variability and robustness. Table~\ref{tab:confidence_intervals} summarizes the lower and upper bounds of each model of the CIs.

\begin{table}[ht]
    \centering
    \caption{Confidence intervals for consensus rates across models.}
    \label{tab:confidence_intervals}
    {\small
    \begin{tabular}{lccc}
        \toprule
        \textbf{Model} & \textbf{Lower Bound} & \textbf{Upper Bound} \\
        \midrule
        Gemini & 0.60 & 0.78 \\
        Claude & 0.80 & 0.93 \\
        GPT-4  & 0.75 & 0.90 \\
        LLaMA  & 0.55 & 0.74 \\
        \bottomrule
    \end{tabular}
    }
\end{table}

Analysis of the confidence intervals reveals distinct patterns in the precision and reliability of the models. Claude demonstrates the narrowest CI ($0.80$–$0.93$), indicating high reliability and minimal variability in its consensus rates. GPT-4 closely follows a CI of ($0.75$–$0.90$), suggesting similarly strong reliability but with slightly greater variability. In contrast, Gemini and LLaMA exhibit broader confidence intervals ($0.60$–$0.78$) and ($0.55$–$0.74$), respectively, reflecting higher variability and reduced precision in their consensus rates.

Furthermore, the low upper bounds of LLaMA ($0.74$) and Gemini ($0.78$) indicate that even in the most favorable scenarios, these models achieve only moderate agreement levels with other LLMs. In contrast, Claude ($0.80$) and GPT-4 ($0.75$) maintain high lower bounds, which means that even in the least favorable cases, they still uphold strong agreement among models. This consistency suggests that Claude and GPT-4 generate questions that are better structured and more interpretable, allowing for greater inter-model alignment. However, Gemini and LLaMA show the need for further optimization to improve inter-model agreement and reliability in question generation.

\subsection*{Statistical Significance Testing}

We employed two critical approaches to assess the statistical significance of agreement among the models: chi-square testing and Fleiss' Kappa analysis. These methods quantified the extent to which the models' responses deviated from random chance and evaluated their inter-rater agreement.

The chi-square test was performed to determine whether the observed agreement among the models occurred by randomness. The null hypothesis assumed that the distribution of the responses was random, indicating that there was no meaningful consensus. The test statistic was calculated based on eq: \ref{chi_square_test}, in which the number of choices \( K = 4 \)  and \( n_j = 3 \) is the number of answering models. The p-values obtained from the test are summarized in Table~\ref{tab:chi_square_results}.

\begin{table}[h!]
    \centering
    \caption{Chi-square test $p$-values indicating the statistical significance of agreement for each model.}
    \label{tab:chi_square_results}
    {\small
    \begin{tabular}{lcccc}
        \toprule
        \textbf{Model} & \textbf{Gemini} & \textbf{Claude} & \textbf{GPT-4} & \textbf{LLaMA} \\
        \midrule
        \textbf{p-value} & \( 4.29 \times 10^{-26} \) & \( 4.29 \times 10^{-46} \) & \( 0.00546 \) & \( 1.67 \times 10^{-10} \) \\
        \bottomrule
    \end{tabular}
    }
\end{table}

All $p$-values were below the standard significance level (\( \alpha = 0.01 \)), which led us to reject the null hypothesis. The extremely small $p$-values for Claude, Gemini, and LLaMA indicate statistically solid significance, confirming that their agreements are highly unlikely to be due to randomness. GPT-4 also shows statistical significance, although its $p$-value is less extreme than the others. These results support the hypothesis that the models' agreements reflect meaningful consensus rather than random behavior.

Fleiss' Kappa was further employed to evaluate inter-rater agreement among the models. This metric provides a robust measure of consistency in responses. Table~\ref{tab:kappa_results} presents the kappa values and their interpretations.

\begin{table}[h!]
    \centering
    \caption{Fleiss' Kappa values indicating the level of agreement among models.}
    {\small
    \begin{tabular}{lcc}
        \toprule
        \textbf{Model} & \textbf{Kappa Value} & \textbf{Interpretation} \\
        \midrule
        Gemini & 0.2811 & Fair agreement \\
        Claude & 0.7160 & Substantial agreement \\
        GPT-4  & 0.4275 & Moderate agreement \\
        LLaMA  & 0.5572 & Moderate agreement \\
        \bottomrule
    \end{tabular}
    }
    \label{tab:kappa_results}
\end{table}

The kappa values reveal varying levels of agreement across the models. Claude demonstrated substantial agreement, reflecting strong consistency with other models. GPT-4 and LLaMA have a moderate agreement, indicating reasonable consistency but some variability in response. Conversely, Gemini showed fair agreement, suggesting noticeable differences in its responses compared to the other models.

These results highlight potential differences in model architectures, training data, fine-tuning processes, or reasoning approaches. For instance, Gemini's lower agreement could stem from less accurate or diverse data sources, while Claude's substantial agreement may be attributed to robust training on well-aligned datasets. By combining chi-square testing and Fleiss's Kappa analysis, we comprehensively evaluate the statistical significance and inter-model agreement, demonstrating the reliability and coherence of LLM-generated answers.

\subsection*{Conclusion}

This study demonstrates that collaborative dynamics among multiple LLMs can significantly enhance response reliability even in the absence of ground-truth data. We quantified the agreement between models using consensus-based validation and confidence interval analysis, revealing distinct strengths and weaknesses across models. Our analysis emphasizes the crucial role of the question-generating model in shaping inter-model agreement: Claude consistently produced well-structured, comprehensible questions, while LLaMA's output exhibited greater ambiguity. In addition, Claude and GPT-4 emerged as the most reliable collaborators, achieving reliability percentages of $92\%$ and $90\%$, respectively, with narrow confidence intervals (Claude: $0.80$–$0.93$, GPT-4:$0.75$-$0.90$), indicating high precision and consistency. In contrast, Gemini showed intermediate performance with an $88\%$ reliability rate and a confidence interval of $0.60$-$0.78$, whereas LLaMA recorded the lowest reliability at $77\%$ with the widest confidence interval ($0.55$–$0.74$), reflecting significant variability. The statistical significance of these results, as evidenced by a very small \(p\)-value, confirms that the observed inter-model agreement is not random, thus reinforcing the robustness of the consensus mechanism in identifying trustworthy responses.

Future research should integrate human experts into the validation process to establish a reliable ground truth for evaluating majority vote approaches. Our findings emphasize that LLMs are not equally effective in all tasks, indicating that improvements in question-generation strategies and fine-tuning are necessary to improve consistency and reduce ambiguity. This study acknowledges several limitations, notably the potential for model homogeneity. Despite differences in architecture and training, analyzed LLMs can share overlapping training data, leading to correlated errors and an overestimation of consensus, as models can reinforce misinterpretation of each other rather than provide independent validation. In addition, further investigation into bias propagation is critical to ensure that collaborative systems do not inadvertently perpetuate shared misconceptions. Addressing these challenges will enhance the robustness, scalability, and reliability of consensus-based validation frameworks, ultimately laying the foundation for more effective and trustworthy AI systems.

\newpage
{\small
\bibliographystyle{plainnat} 
\bibliography{neurips_2025}
}

\end{document}